# Hyperparameters optimization for Deep Learning based emotion prediction for Human Robot Interaction

Shruti Jaiswal, and Gora Chand Nandi, *Member, IEEE*

**Abstract**—To enable humanoid robots to share our social space we need to develop technology for easy interaction with the robots using multiple modes such as speech, gestures and share our emotions with them. We have targeted this research towards addressing the core issue of emotion recognition problem which would require less computation resources and much lesser number of network hyperparameters which will be more adaptive to be computed on low resourced social robots for real time communication. More specifically, here we have proposed an Inception module based Convolutional Neural Network Architecture which has achieved improved accuracy of upto 6% improvement over the existing network architecture for emotion classification when combinedly tested over multiple datasets when tried over humanoid robots in real - time. Our proposed model is reducing the trainable Hyperparameters to an extent of 94% as compared to vanilla CNN model which clearly indicates that it can be used in real time based application such as human robot interaction. Rigorous experiments have been performed to validate our methodology which is sufficiently robust and could achieve high level of accuracy. Finally, the model is implemented in a humanoid robot, NAO in real time and robustness of the model is evaluated.

**Index Terms**—Emotion Classification, Optimized Deel Learning, Convolutional Neural Network, Inception Module, Human Robot Interaction.

———————————— ◆ ————————————

## 1 INTRODUCTION

Robots have now entered human life. An era has reached where human beings have got a helping hand such that human just needs to give direction to robots and get the work done. In such a scenario, it will be an add-on for us if robots also understand our emotion. Human face is a complex structure which can be manipulative at times. Predicting facial expression by looking at the image can be complex even for human. Facial expressions depends on many factors, like, the context in which the person is talking, the sentiment of the statement person is speaking or listening to, and the entire situation of the scenario. Hence all these factors makes predicting emotion just by looking at the image could be more difficult to achieve accuracy.

When a robot is capable to predict human emotion, it can enter in a person's life in a more comfortable manner that means it can come up with collaborative behavior with human. It would be easier for a personso connect to a robot. Then robots can be customized as per person's need which can be useful in conducting therapies for child, taking care of elderly peopleor even living with person as a daily routine.

When it comes to human facial expression by looking at an image, human performance itself has achieved 65± 5 % accuracy [1] that was researched over Fer2013[2] dataset. This was computed by manually predicting expression from every image of the dataset.

Problem of making the robot predict human emotion, in order to stay with them in a collaborative manner, many ways have been achieved to solve it. Initially, this complex problem was solved in an early stage in year 2000-2004 when researchers in [3] manually extracted out features of the emotion dataset images so that when fed to neural network, it can predict the emotion. Much research has been done on feature extraction since then like extracting upper face action unit, lower face action units, including eyebrows, lips, check structure, and chin lifted, eyes shape and many others. Then these these features are used to predict human emotion. Even neurophysical data is also used to predict human emotion as in [4]. In addition to that, deep learning has replaced the process of manually extracting features or manually extracting action units; one such work is done in [3]. Facial action recognition is done similar grounds using static face images in [5]. In this paper, we have proposed a model to predict human emotion in a more optimized and accurate manner.

Detecting human emotion is still a crucial task to solve and also complex to make it real time implementable. To take up this challenge, in this paper, we have proposed a model to make emotion detection more accurate over many variations of the dataset and also more real times implementable with less time complexity over existing models and hence have implemented a humanoid robot NAO to be able to predict human emotion while it is talking to the person. Main contributions of this paper are:

————————————————
- *Shruti Jaiswal is with the Indian Institute of Information Technology, Allahabad, Jhalwa Allahabad, UP-211015, India. E.Mail: shruti.jaiswal123@gmail.com.(+91-9717105219)*
- *G.C. Nandi is Senior Most Professor with the Department of Information Technology, Indian Institute of Information Technology, Allahabad, Jhalwa Allahabad,UP-211015, India. E-mail: gcnandi@iiita.ac.in.*

- Building robust model which has been tested over 8 datasets.
- Reduction in time complexity of the model.
- Time complexity analysis is done to justify that model is implementable in real time.
- Response time and accuracy analysis of the model has been performed on a real humanoid robot platform, NAO.

Further, the paper is structured with analysis of previous research for this application in the following section. After that in section 3 problem formulation is explained. Section 4 has proposed methodlogy, it explains Inception module, feature extraction, facial emotion recognition details. Here, we have also discussed Implementation feasibility of our proposed model on humanoid robot. In section 5, we have described details of all the dataset used to generalize the model. Finally section 6 has experiment details and result analysis. And lastly in section 7 and 8, we have summarized our paper with conclusion and future work of our proposed method respectively.

## 2 ANALYSIS OF PREVIOUS RESEARCH

Lots of research is done in past few years, starting from basic neural network to predict human emotion. Researchers in [6] have used CNN to detect Action Units (AUs) for the faces in the emotion dataset to predict target emotion. This method extracts AUs from image sequence to predict final emotion state. This processing via CNN over sequence of images increases the complexity of proposed model.Authors in [7] have built a probabilistic model to generate human emotion on android robots using AUs. Researcher in [8] have used Support Vector Machine (SVM) to classify seven different emotion using Gabor wavelet. This method is still complex in implementation on a real-time system. To optimize complexity of a CNN for image classification task, authors in [9] have build a classifer using Principal Component Analysis (PCA), to reduce the dimensionality and then CNN to extract the features. Such a model is good with reduced complexity for image classifer applications, but when it comes to detecting emotion, which is highly vibrant in nature, reducing dimensionality at this early stage will lead to reduction in efficiency. In [10] also, authors have built a model to predict human emotion using PCA + Artificial Neural Network (ANN). Using ANN for image data would enhance parameter requirement and hence increased complexity. When analysed over L1 and L2 regularization to extract image features is done in [11]. They have shown which regularization to use in what kind of application requirement differentiating in extracting local features and global features. Shima et. al. in [12] have tried to reduce the overfitting of the model by introducing dropout, batch normalization and L2 regularization in CNN and have verified the feature extraction gives the same effect when extracted only by CNN or when Histogram of Oriented Gradients (HOG) features are used in parallel with CNN.

Researchers in [13], have improved on AlexNet performace to predict human emotion by varying the number of convolution layers as per the application need. In [14], authors have added SVM for classification purposes of features extracted via AlexNet model. Representational autoencoders and CNN model is built in [15] to predict human emotion.

In recent years, efforts are done to optimize and improve efficiency for emotion prediction using deep learning techniques. Authors in [16] have used CNN to extract out features and then implemented SVM for classification over translation invariant features and not over hand crafted features as used in earlier research. Recently, EEG dataset is also used to predict human emotion in [17] using DEAP dataset. Here, deep neural network and CNN model are built and analysed over EEG data for emotion prediction rather than facial images. Further in [18] paper, authors have built a model using genetic algorithm (GA) + SVM for EEG signals dataset. They have done the classification on two dimensional emotional model, arousal and valence. Electrodemal Activity [EDA] signals are also used for emotion classification as in [19]. Classification here is done using SVM classifier. An overall review about emotion classifiers, its accuracy achieved over various datasets and complexity issues are discussed in [20]. This review focusses on various hybrid deep learning approaches for emotion classification problem involving CNN, Recurrent Neural Network (RNN) especially LSTM to capture sequential frames of images to predict emotion.

Later in [21], researchers have incorporated facial landmarks to loss function of classification process. Using facial landmarks helped differentiate between local and global features to help classify the variation in emotion. Action Units are also predicted for emotion classification. In [22], AUs prediction is automated by extracting spatial and temporal representations from the data. Spatial representations are extracted using CNN model and temporal represenations are computed using Long Short Term Memory (LSTM) model. Here, LSTMs are stacked on top of CNN to extract temporal representation after pulling spatial representations. Then combination of AUs brings out the emotion of the person. The network based on CNN proposed by [23] worked to make the network more generalizable by introducing more non-linearity to the model by adding four inception layers to the model. This contribution helped in generalizing the model to a better extent and can cover up more unknown faces, or unknown variant of faces with emotion. CNN is tried and improved by different ways depending on the requirement and application by working upon layers, regularization, loss function, activation function and computation speed. Such analysis is done in [24]. In [25], authors have further tried to improve upon the network by adding non-linearity in the system and worked upon the complexity by added global average pooling in place of fully connected layers.

## 3 PROBLEM FORMULATION

To solve the real time implementational need for a social personal robot in order to predict human emotion for collaborative behavior, we have proposed a CNN model based on Inception modeule concept and have verified its real time complexity. Requirement of hyperparameters to be trained in a deep learning model is huge resulting in requirement of huge computational resources. Hence main contribution of this research is optimized model to predict human emotion which is robust in nature (has been verified over seven datasets) .This model is implemented over humanoid robot NAO and we have tested its response time and true predictions percentage.

A comparison is done between state of the art existing models and our proposed model on metrics of number of parameters, number of weighted layers, training accuracy percentage and test accuracy percentage.Also the model is tested over seven different datasets to validate its robustness. A comparison is done in response time of the robot to predict human emotion while the robot if talking to the person.

## 4 PROPOSED METHODOLOGY

Our proposed model is based on Inception module, and is evaluated on criteria of accuracy, generalizability, response time in humanoid robot and computational speed while training and testing of the model. Inception module is used to introduce non-linearity to the system and reduce dimension as and when required. We have targeted to reduce dimension gradually down the network as we go deep, so as to not miss any feature in the beginning. Architecture of the network is shown in Fig. 1. We have tested our proposed model on seven different datasets namely, Fer2013, JAFFE, CK dataset, CFD, Impa-face3D dataset, Affectnet and a custom dataset built in our lab. It is tested over seven class labels namely, "happy", "sad", "angry", "neutral", "surprise", "fear" and "disgust".

To optimize the model, we have worked on batch normalization [26], 1×1 convolution [26], depthwise separable convolutions and global average pooling (GAP) [26]. The use of Gap reduces the parameter requirement to a great extent by approximately 85%. Finally the network is tested over humanoid robot, NAO along with vanilla CNN model with almost same number of layers and performances of both situations are compared and analyzed.

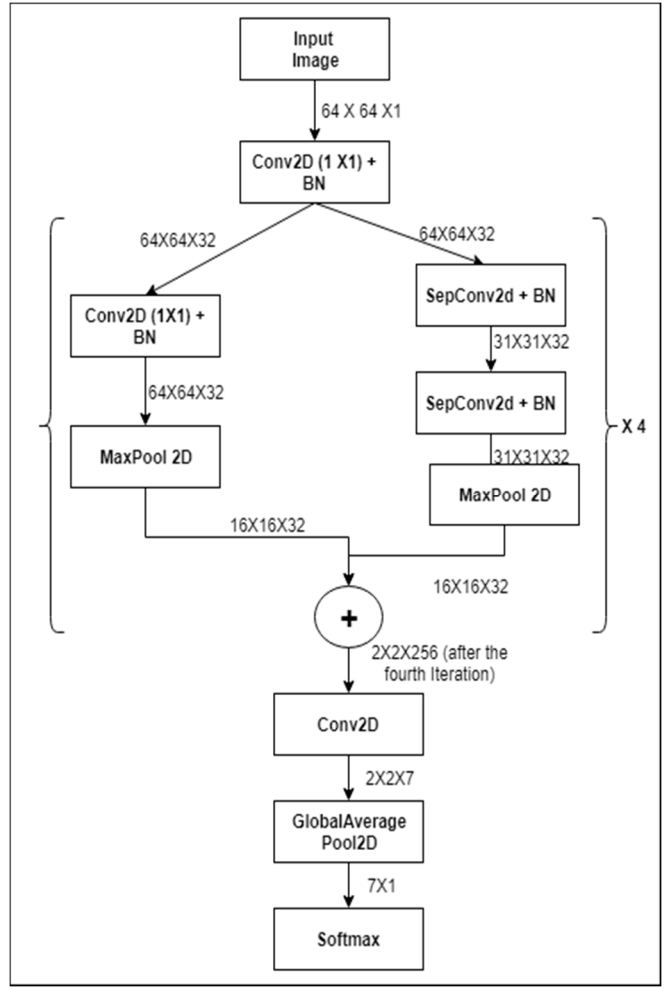

Fig. 1. Architecture of our proposed model. (filter size is increased in the loop from 3 to 5 to 7 and then to 9 to gradually lower the dimension thereby retaining the best feature for classification).

### 4.1 Inception Module
Inception module [26] helps reduce the complexity as it uses 1×1 kernal size before using any bigger size kernel and also reduces channels thereby reducing the computation. Also, adding 1×1 filter in parallel with other filters help extract other variants in the features which higher size filter convolution misses out. These two benefits are utilised in our model along with few other concept which will be explained subsequently to make it computation efficient and still be able to extract out the best feature set and meet state of the art accuracy.

### 4.2 Feature Extraction
Extracting features from face dataset to predict the emotion is a crucial task. It is difficult to achieve or improve accuracy in the sense that slight variation in the angle of lips, tilted eyebrows etc can result in change in the prediction of class or emotion in our case and lead to misclassification. We have increased the filter size gradually in the network as we go deep, so that dimensionality reduces with depth so that we don't miss out relevant feature in the beginning.

### 4.3 Facial Emotion Recognition
The proposed network architecture is shown in Fig. 1. In the first layer, Convolution 1X1 is used to reduce dimensionality and computation and also introduce more non linearity in the system. After that, separable convolution is implemented 2 times in a loop of 4 with max pool operation as shown in architecture. Along with separable convolution, 1X1 convolution is added with the result as residue as shown in the architecture. This is done considering any missing feature that can be brought up from previous layers.

  In the architecture, a part is repeated four times with increasing number of filters used each time. Number of filters for convolution is increased from 32 to 64 to 128 and finally to 256. Depth in the feature map is increased gradually so that best of the features can be extracted. Dimensionality is reduced gradually so that relevant feature is not missing with increasing convolution operations.

### 4.3.1 Parameter Reduction Strategy

We have introduced many strategies to reduce parameter requirement in our model. In the beginning layer, 1X1 convolution is used to reduce the dimension of the input image and introduce non linearity in the dataset for better generalization of the network. Further separable convolution is used which reduces the number of multiplications in every epoch.

After feature extraction in Convolution layers, classification task is performed. For classification, instead of Fully Connected layers, we have used Global Average Pooling which globally extracts out the average of the features. So average of the best features is highlighted out to perform the classification using Softmax function.

### 4.3.2 Reduction in Computational Resources

Use of separable convolution reduced the number of multiplication required in every epoch by a factor of 15 (approx.) as compared to convolution operation. Batch normalization is used which reduces the range of values hence introducing simpler multiplications. Also use of Global Average Pool instead of fully connected layers reduces all the computation of training the parameters of the dense layers. Global Average Pool performs average pooling operation over final set of feature map which only involves very few multiplications. In the last convolution layers, we have generated 7 channels of feature map as we have 7 classes. Global average pooling averages the 7 feature map and Softmax then classifies it to the most dominant emotion.

### 4.3.3 Performance and Accuracy Trade off

Use of separable convolution reduced the number of multiplication required in every epoch by a factor of 15 (approx.) as compared to convolution operation. Batch normalization is used which reduces the range of values hence introducing simpler multiplications. Also use of Global Average Pool instead of fully connected layers reduces all the computation of training the parameters of the dense layers. Global Average Pool performs average pooling operation over final set of feature map which only involves very few multiplications. In the last convolution layers, we have generated 7 channels of feature map as we have 7 classes. Global average pooling averages the 7 feature map and Softmax then classifies it to the most dominant emotion.

### 4.4 Implementation to check its real time applicability on Humanoid Robot, NAO

Personal Robots can be made more customizable when it understands the human emotion along with conversation and visualization. It will help train/ treat elderly, child or other needy person to get their personal work done in a better way instead of robots becoming annoying for human. For our experiment, we have used NAO Humanoid Robot, though our application is not limited to NAO or to humanoid robots. NAO is a humanoid robot which is mobile, agile and Interactive. It has 25 degree of freedom. For vision, NAO has 2 cameras, one on the forehead center and other on the lips. We have used forehead camera of NAO to make it predict emotion of the person to whom NAO is looking at.

## 5 VARIATIONS IN THE DATASET

Datasets used for training, validation and testing our model is FER2013 dataset which comprises of 28709 training samples from 7 class labels and 3589 images for testing the network. Labels in the dataset are 0=Angry, 1=Disgust, 2=Fear, 3=Happy, 4=Sad, 5=Surprise, 6=Neutral. Testing is done on different datasets to verify the robustness of the model. For testing the network, JAFFE dataset [27] and custom dataset are used. JAFFE dataset contains 230 images of 10 Japanese female models all posed for 7 different emotions. Custom dataset consists of 2000 images. Out of which 840 images are posed for 7 emotions, 700 images are collected with spontaneous emotions by showing movie clips, and collected during different moments of people gathering at different occasions. 300 collected from internet with different emotion labels, and 160 images are non-face images for adding generality in the dataset.

Many variants in the dataset is used like different light conditions, face makeup, Asian and non-Asian faces, Japanese face, different background, all age groups and even non face images in order to achieve better generalization. All the datasets used for the experiment are described below showing number of test and train images for each class label. And it shows the subject count considered for dataset telling that inclusion of all these dataset assured addition of every variation to the set thereby network should be able to handle every kind of data well, instead of performing best for any one particular type of dataset and failing for others.

### 5.1 Fer2013 dataset

As explained earlier, this dataset is used for training and testing of our model. This dataset has 32,297 images in total and is divided in 28709 train and 3589 test images. We have verified manually that this data consists of non face images as well to embrace genrality in the network.

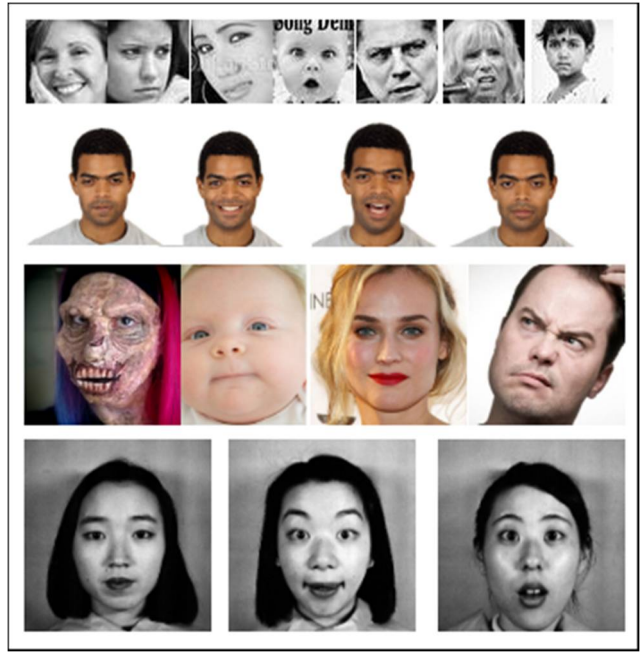

Fig. 2. Image samples of some datasets. First row has samples of Fer2013 dataset, second row has CFD dataset, third row is sample from AffectNet and finally samples of JAFFE dataset are shown.

### 5.2 JAFFE

JAFFE is quite old small dataset for emotions consisting of 213 images of 7 emotion labels. We have considered this also for testing as it is used by many researchers to create a benchmark of results, hence comparitive confusion matrix is also build over this dataset.

### 5.3 CK dataset [28]

This dataset is relevant for our experiment in the sense that it has consecutive frames of images to predict an emotion. We are also performing similar experiment while implementing our model in humanoid robot, NAO, which captures 10 frames per second to predict human emotions.

### 5.4 CFD [29]

This is Chicago Face dataset, it has 1250 images of categories: neutral, happy (with open face), happy (with close face), angry and fearful. We have verified this dataset using 4 labels: happy, neutral, angry and fear.

### 5.5 IMPA-FACE3D dataset [30]

This dataset consists of 266 images from 38 subjects of 7 emotions which we are considering in our experiment. When our model is tested over this dataset, it has achieved 77% of accuracy.

### 5.6 AffectNet [31]

Affectnet dataset has 420,299 images of 11 categories namely, neutral, happy, sad, surprise, fear, disgust, anger, contempt, none, uncertain, and non-face. Out of which we are labeling it to first 7 categories and rest are considered under other category in our case.

### 5.7 Custom Dataset created in our lab

We have created a dataset of 2000 images consisting of posed and spontaneous images of 7 different emotions using 12 subjects and further collected group face images of the subjects to classify emotions in them and test our model.

## 6 EXPERIMENT DETAILS AND RESULTS ANALYSIS

We have used seven different datasets for our processing as discussed earlier. This experiment testing on various datasets is done to ensure robustness of the network and to check if that network can generalize to all the variation in the face dataset whether it be light effects, face changes in different region of the world, angle of the face and many other variations. After proof validation of our model, we have tested it on humanoid robot, NAO. We have also tested vanilla CNN model for comparison and to show real time effectiveness of our model over basic model.

The architecture for the model is shown in the fig. 1. The model is based on inception module [26], involving 1X1 Con-volutions and Global Average Pooling to reduce the number of trainable hyperparameters. In this network, separable Convolutions are used to reduce the hyperparameter requirement. 1X1 Convolutions are used to further reduce complexity and fully connected layer is replaced with global average pooling to further reduce most of the trainable parameters in order to make it real time implementable. Few samples of the datasets used for training and testing from various datasets mentioned above are shown in Fig. 2. In that figure, first row has samples of Fer2013 dataset, second row has CFD dataset, third row is sample from AffectNet and finally samples of JAFFE dataset are shown.

Fig. 3 shows the custom data created in our lab consisting of 2000 images used for testing purposes. Rigorous training and testing has been performed to verify the validity of the trained model. Images were taken in different background and light conditions. And mostly all the seven emotions were captured for different person. Label was created with manual judgement which also might have discrepancy, as human accuracy to judge an emotion is 65±5%. Some unlabeled data was also generated to test our network, like when face is not too clear with light conditions, or non face image shown in Fig. 3. No class data images are added in the dataset for further generalization of the network.

The network was trained on GPU server in which training was done several times in order to freeze the hyperparameters and the layer structure which can provide most efficient result with least computational cost. As the Occam Razor's principle, the simplest algorithm solving the same problem is the best one. As the fully connected layer in a CNN model has mostly 90 % of the parameters of the entire network, so using Global Average Pooling [26] has reduced the excess parameters. Further parameters are more reduced by adding 1X1 convolutions which can drastically reduce the parameter requirement and using separable convolutions which further reduces the parameters to be trained. Using depth-wise separable convolutions further reduces parameter. It works as convolving as two sequential layers (on the name of one Convolution), namely depth-wise convolution and point-wise convolution as shown in Fig 3. Point-wise convolution is basically 1X1 convolution in order to reduce the parameter requirement. Say if we have input of 16 channels and Convolution operation is performed using 32 3X3 filters, in that case all the 16 channels are traversed by every 32 channels of 3X3 size, so the number of parameters are:

16 X 32 X 3X 3 = 4608

While when separable convolution is used, it traverses 16 channels by 1 3X3 filter and also 16 channels by 32 1X1 filters, thereby resultant number of parameters becomes:

(16 X 1 X 3X3) + (16 X 32 X 1X 1) = 656

Hence the number of required parameters reduces to a great extent performing the same task achieving desired accuracy.

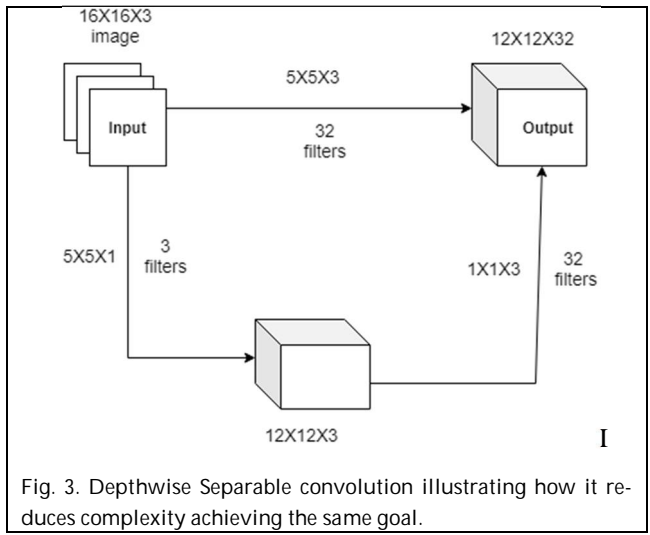

Fig. 3. Depthwise Separable convolution illustrating how it reduces complexity achieving the same goal.

## 6.1 Experimental Setup with Humanoid Robot, NAO

After intense training and testing on all the possible variations of input, the network is implemented on NAO humanoid robot. NAO forehead camera is used as input for the network. The maximum number of predictions out of 10 frames per minute of NAO vision is predicted as final emotion of the person to which NAO is talking to. This has been repeated every 2 minutes during the experiment. 15 subjects were used in the experiment to talk to NAO who tried making all the 7 emotions after every 2 minutes while talking to NAO as shown in Fig. 4. Each subject made the emotions two times to evaluate the efficiency of the system.

The efficiency in the experiment is increased as compared to that in the network as here maximum prediction is considered out of 10 consequent predictions, leading to correct prediction most of the time.

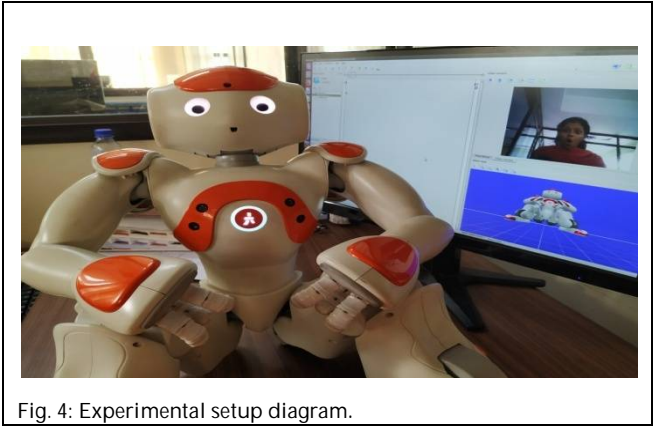

Fig. 4: Experimental setup diagram.

## 6.2 Experimental Result and Analysis with extracted features, complexity & accuracy

After evaluating the model over various datasets, the accuracy achieved on Fer2013 dataset is 72% whose plot is shown in the Fig. 5(a). Fig. 5(b) shows the decrease in the loss showing improvement in the performance of the training procedure. This accuracy and loss are for the training of the model.

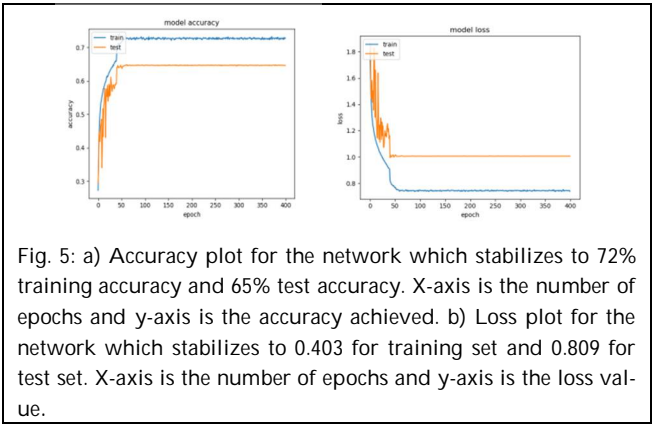

Fig. 5: a) Accuracy plot for the network which stabilizes to 72% training accuracy and 65% test accuracy. X-axis is the number of epochs and y-axis is the accuracy achieved. b) Loss plot for the network which stabilizes to 0.403 for training set and 0.809 for test set. X-axis is the number of epochs and y-axis is the loss value.

Fig. 6 shows the results obtained in our custom dataset build in our lab. Its accuracy is approximately 64% - 68%. The network built is able to extract out faces from group photo and predict emotion from it as shown in the figure. For face detection in our model we have used haarcascade_frontalface model designed by OpenCV which detects the frontal face. This algorithm is based on Viola Janes detection algorithm trained on face and non-faceimages to identify a face. Clearly effiecieny of our model is also affected by efficiency achieved in the face detection model.

Comparison on confusion matrix for JAFFE dataset is shown in Table 1, and 2 of Fig. 7 for Vanilla CNN, and our model respectively. These tables show that Vanilla CNN (a variant of AlexNet) lagged in prediction of angry and disgust emotion at all and also percentage accuracy for other emotions were also lesser as compared to our model.

Confusion matrix for CFD dataset is shown in Fig. 8. This dataset has 4 classes namely, neutral, angry, happy, fear and some non class images. When tested on our network, other three labels: surprise, sad, and disgust if predicted are considered as non-class as these doesnot exists in the dataset. So confusion matrix is framed accordingly. It is achieving 74.45 % accuracy as shown in the disgram. We have also tested our model on AffectNet dataset. In this, we have taken random 700 images from manually annotated validation set and achieved 65% accuracy as shown in confusion matrix in Fig. 9. Affectnet dataset has 10 classes consisting of contempt, none, uncertain and non-face images which are additional than 7 emotions considered by us. So all other additional emotion if predicted or falsely predicted is considered in class "other" as shown in the diagram.

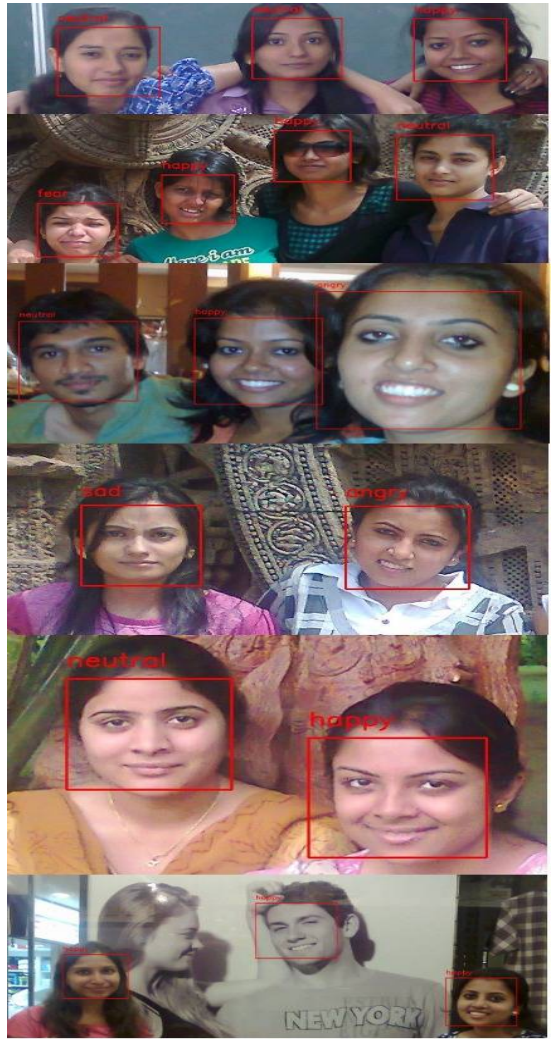

Fig. 6. Result of emotion classifier obtained from all the datasets used for verification.

The efficiency obtained on NAO is further improved from that predicted by the trained model, as the maximum predicted emotion is finalized out of 10 frames in a minute. Results of happy and sad emotion predicted by the NAO robot is shown in Fig. 10, where the robot is trying to initiate a conversation with human based on the predicted emotion in the choreographe simulation software. Here NAOQi api is used for speech generation of the robot and to capture images for emotion prediction using the NAO forehead camera.

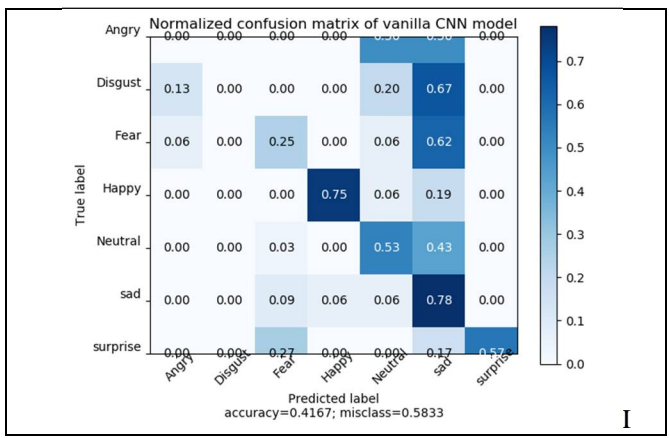

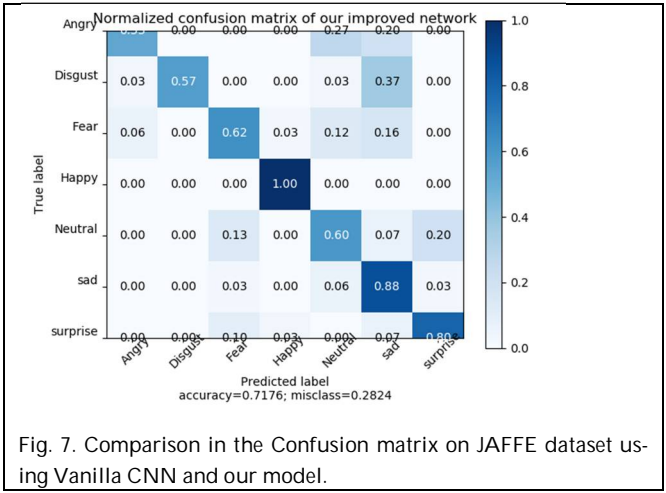

Fig. 7. Comparison in the Confusion matrix on JAFFE dataset using Vanilla CNN and our model.

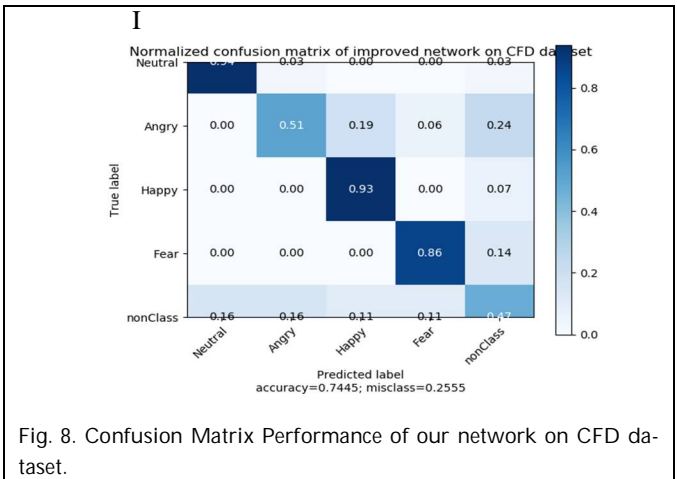

Fig. 8. Confusion Matrix Performance of our network on CFD dataset.

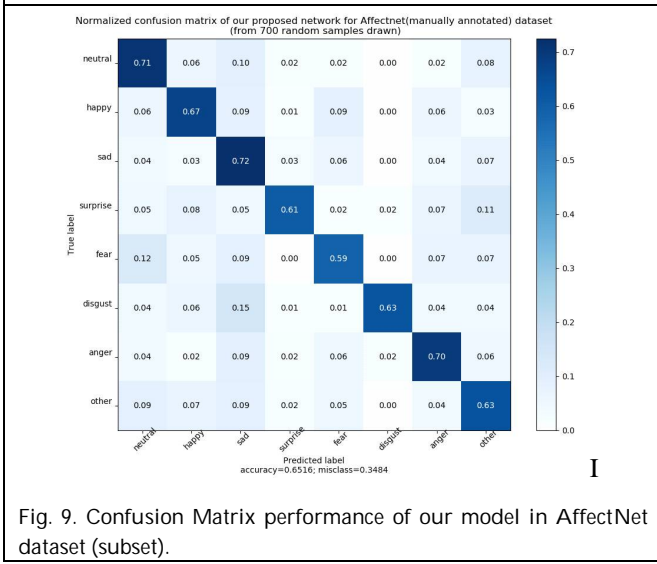

Fig. 9. Confusion Matrix performance of our model in AffectNet dataset (subset).

## 6.3 Reduction in Complexity and its real time implementation ability

Here we are comparing the time complexity of vanilla CNN model, and our proposed model. To create similar conditions for comparison, both the models were created with global average pooling for classification purpose hence we donnot need to compare fully connected layer parameters. Also for CNN, time complexity of the model computed using [32] is:

$$O(\sum_{l=1}^{d} n_{l-1} \cdot s_l^2 \cdot n_l \cdot m_l^2) \qquad (1)$$

where, $n_l$ is number of filters in $l^{th}$ layer

$s_l$ is size of the filter i.e. 5 for 5×5 filter

$m_l$ is ouput feature map spatial size for layer l.

For simplicity in computation, we have considered only the theoretical complexity and hence $m_l$ can be omitted. Hence, our equation for computation becomes:

$$\sum_{l=1}^{d} n_{l-1} \cdot s_l^2 \cdot n_l \qquad (2)$$

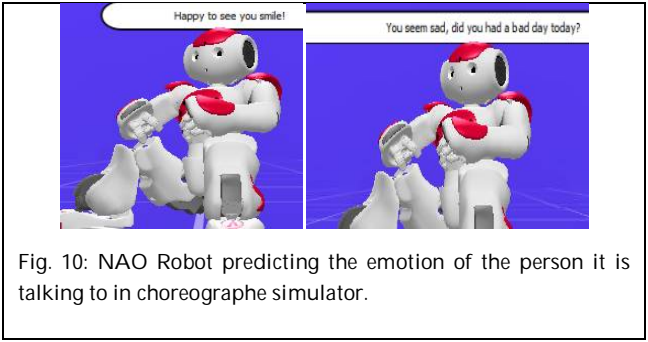

Fig. 10: NAO Robot predicting the emotion of the person it is talking to in choreographe simulator.

As stated in [32], this computed time complexity applies to both training and testing. Training time is roughly three times that of test time per image. Testing an image needs one forward propogation of the model, where as training one image needs one forward and twice backward propagation as described in [32].

Also use of Depthwise Separable Convolution in our model rather than Convolution reduces the number of multiplication as explained in fig 3. Use of seperable convolution reduces the computation by a large amount (8 times approximately lowered in our example explained earlier), which futher adds to the fact that our model is lowered in computation as compared to other models.

### 6.3.1 Time complexity of Vanilla CNN model

The Vanilla CNN model is framed with 12 layers of Convolution operation with batch normalization and ReLU activation function. The numbers of layers match with that of our model in order to maintain the equality to compare. The fully connected layers are replaced with global average pooling so that all we need to compare is just convolution operation and its complexity. Using layers input and output number and size of filters its time complexity can be computed as:

$T(n) = 16 \times 5^2 \times 16 + 16 \times 5^2 \times 32 + 16 \times 5^2 \times 32 + +16 \times 5^2 \times 32 + 16 \times 5^2 \times 32 + 16 \times 5^2 \times 32 + 16 \times 5^2 \times 32 + 16 \times 5^2 \times 32 + 16 \times 5^2 \times 32 + 16 \times 5^2 \times 32 + 16 \times 5^2 \times 32$

$\qquad = 1{,}144{,}320 \qquad (3)$

### 6.3.2 Time Complexity of our proposed model

Our proposed model, architecture explained in figure 1, having almost same number of convolution operations with global average pool at the end, having time compexity computed as:

$T(n) = 32 \times 1^2 \times 32 + (32 \times 1^2 \times 64 + 64 \times 1^2 \times 128 + 128 \times 1^2 \times 256 + 256 \times 1^2 \times 7) + (32 \times 3^2 \times 64 + 64 \times 3^2 \times 128 + 128 \times 3^2 \times 256 + 256 \times 3^2 \times 7) + 7 \times 5^2 \times 7$

$\qquad = 450{,}249 \qquad (4)$

Clearly, as compared with equation (3), complexity is almost reduced to one third of its value as shown in (4) and hence implementable in real time systems which we have explored subsequently.

### 6.3.3 Real Time Implementation on humanoid robot, NAO

The real time experiment setup is shown in fig 4. Here you can see the subject is sitting in front of the robot and talking to it. Here only human emotion prediction using face image is used, hence whatever emotion is posed by the person, robot tries to start the conversation with such mood in order to make it feel more natural to connect and communicate. Response time is computed for each emotion on different subjects and has been compared with Vanilla CNN model response time when implemented in NAO robot. Accuracy in prediction and response time is compared for Vanilla CNN and our model and comparison table is shown in Table 1. Here we have obtained minimum of 0.45 seconds and maximum of 1.02 seconds on an average for all 7 emotions when experimented over 10 subjects. Whereas when Vanilla CNN is used for the same experiment, almost all the emotions were predicted with longer delay and prediction rate is also reduced. This response time and prediction rate was computed by making the person sit in front of the robot while holding that emotion till Robot predicts it (right or wrong) with a threshold of 1.30 seconds.

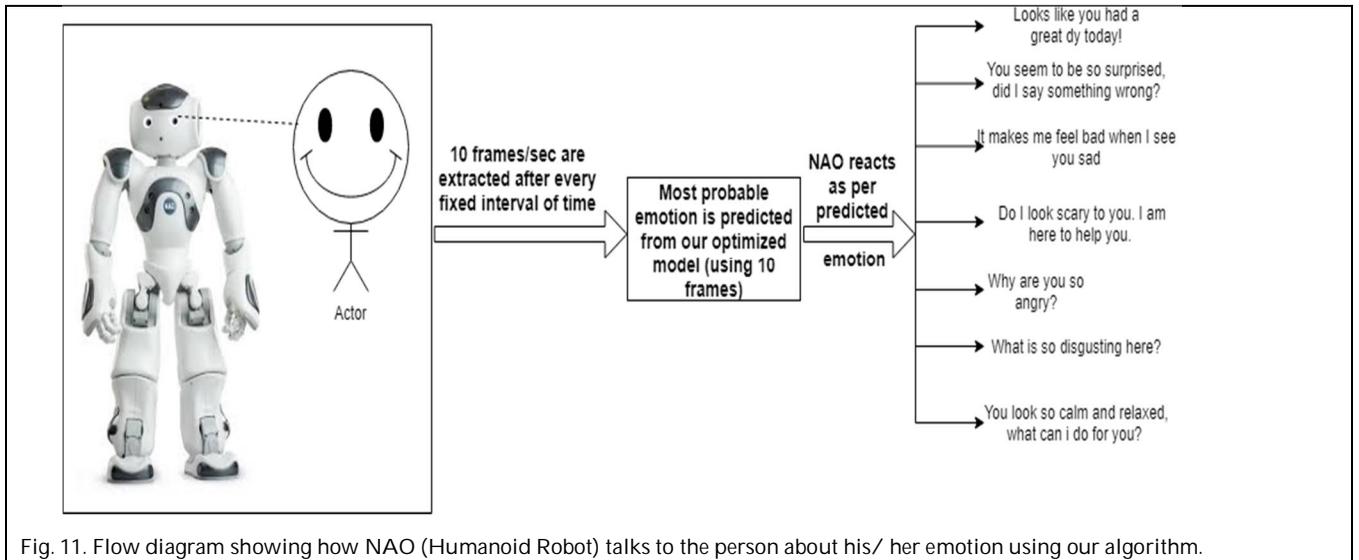

Fig. 11. Flow diagram showing how NAO (Humanoid Robot) talks to the person about his/ her emotion using our algorithm.

Flow diagram on how the experiment was performed to compute response time and prediction rate is shown in Fig. 11. During the experiment, 10 frames per second was captuered after every 2 minutes and if any emotion is dominating in those frames, NAO speaks out about the emotion to check the well being of the person it is talking to. This can be used in many applications like personal robots, treating elderly people, treating autistic patients, physchological treatments of kids and elderly people etc.

Table 1: Average Response Time and prediction rate (using 10 subjects) on NAO when Vanilla CNN and our model are implemented for Prediction.

|  | Vanilla CNN | | Proposed Model | |
| --- | --- | --- | --- | --- |
|  | Response time (sec) | Prediction Rate | Response Time | Prediction Rate |
| Happy | 0.98 | 7 | **0.45** | 10 |
| Sad | 1.03 | 7 | 0.49 | 8 |
| Neutral | 1.26 | 4 | 0.72 | 6 |
| Angry | NA | 0 | 0.86 | 7 |
| Surprise | 1.29 | 8 | 0.63 | 8 |
| Fear | 0.96 | 3 | **1.02** | 6 |
| Disgust | 1.58 | 1 | 0.91 | 6 |

## 7 CONCLUSION

In order to understand human's sentiment in a statement, robot need to be able to understand sentiments in text, but to relate to the context in which it is being said, understanding facial expression will also help to a great extent. Hence to make humanoid robots understand people in a more efficient manner, we have made an effort such that facial expression can be estimated in a real time manner. This will make robots connect to human in a better way, and human can rely or get more comfortable in presence of robots. Even this way human can express themselves to robots and stay more personalized.

Our proposed model is tried, tested with humanoid robot and improved responses are shown. Our network achieved 94% reduction in parameter requirement, leading to reduction in complexity and an overall improvement in accuracy of upto 6% is reached when implemented over humanoid robotswhen compared over latency and response time.We have shown the improvement in real time system, i. e., humanoid robot, NAO.

## 8 FUTURE WORK

Our proposed network once embedded with humanoid robots can judge person's emotion and deal the situation accordingly. Further multi modal communication can be added in robot to make it more social and understandable. Affectnet dataset had further more classes like content, no-emotion, no face, etc. which can be further incorporated in the model. Sentiment of a same statement might be different at different times and situation, which can be predicted clearly if facial emotion is also known at that instance. Also researchers in [33] have mentioned various mixed emotions like "angrily disgusted", "sadly angry", etc, we are looking forward to add these too so that clearer prediction can be made for a better collaborative behavior.

Our future work is directed to improve our models performance on humanoid robot NAO. Two camera are embedded in NAO robot one on the forehead and other in position of the mouth. These can take images in a range of resolution from 160 X 120 up to 1280 X 960. NAOQi framework described in [34], handles the vision api which can be used for both the cameras to improve the performance and try to extract out the best features from both the images of same instance. Authors in [35] have predicted finger pointing direction using single RGB camera. As our robot is already integrated with emotion classifier, adding gesture and speech to robot will make it more smooth and genuine. Hence personal robots can work in applications like psychological counseling, kids handling, elderly care etc. with more ease.

## ACKNOWLEDGMENT

The authors thank all the research scholars of robotics and machine intelligence laboratory of our institute who gave their consent and helped in data collection and carrying out the experiment. This research was improved by the suggestions given by reviewers of CVPR conference where a part of this work is presented as a poster in Women in Computer Vision Workshop of CVPR 2019 Conference.